\documentclass[letterpaper]{article} 
\usepackage{aaai24}  
\usepackage{times}  
\usepackage{helvet}  
\usepackage{courier}  
\usepackage[hyphens]{url}  
\usepackage{graphicx} 
\usepackage{natbib}  
\usepackage{caption} 
\usepackage{algorithm}
\usepackage{algorithmic}
\usepackage{amssymb}
\usepackage{amsmath}
\usepackage{amsmath, bm}
\usepackage{booktabs}
\usepackage{multirow}

\usepackage{newfloat}
\usepackage{listings}
\DeclareCaptionStyle{ruled}{labelfont=normalfont,labelsep=colon,strut=off} 
\floatstyle{ruled}
\newfloat{listing}{tb}{lst}{}
\floatname{listing}{Listing}
\title{Towards More Faithful Natural Language Explanation Using Multi-Level Contrastive Learning in VQA}
\author {
    Chengen Lai,
    Shengli Song,
    Shiqi Meng,
    Jingyang Li,
    Sitong Yan,
    Guangneng Hu
}
\affiliations {
    School of Computer Science and Technology, Xidian University, Xi’an, China\\
    \{laice, ShiqiMeng, jylee, styan\}@stu.xidian.edu.cn,  shlsong@xidian.edu.cn, njuhgn@gmail.com
}

\usepackage{bibentry}

\begin{document}

\maketitle

\begin{abstract}
Natural language explanation in visual question answer (VQA-NLE) aims to explain the decision-making process of models by generating natural language sentences to increase users' trust in the black-box systems. Existing post-hoc methods have achieved significant progress in obtaining a plausible explanation. However, such post-hoc explanations are not always aligned with human logical inference, suffering from the issues on: 1) Deductive unsatisfiability, the generated explanations do not logically lead to the answer; 2) Factual inconsistency, the model falsifies its counterfactual explanation for answers without considering the facts in images; and 3) Semantic perturbation insensitivity, the model can not recognize the semantic changes caused by small perturbations. These problems reduce the faithfulness of explanations generated by models. To address the above issues, we propose a novel self-supervised \textbf{M}ulti-level \textbf{C}ontrastive \textbf{L}earning based natural language \textbf{E}xplanation model (MCLE) for VQA  with semantic-level, image-level, and instance-level factual and counterfactual samples. MCLE extracts discriminative features and aligns the feature spaces from explanations with visual question and answer to generate more consistent explanations. We conduct extensive experiments, ablation analysis, and case study to demonstrate the effectiveness of our method on two VQA-NLE benchmarks.
\end{abstract}

\section{Introduction}

Deep neural networks have achieved significant progress on visual question answering (VQA)~\cite{antol2015vqa}. However, most of them are black-box systems, which makes it hard to gain users' trust. It is a critical problem for these models to explain their decision-making process. In recent years, there has been some development of explainable VQA systems~\cite{patro2019u,chen2022rex}. Visualization analytic approaches generate a heatmap with different values by exploiting attention mechanisms and gradient analysis~\cite{lu2016hierarchical,selvaraju2017grad} where the regions with higher values contribute the most to the predicted answers. However, such visualizations do not explain how these regions support the answer.

\begin{figure}[h]
	\centering
	\includegraphics[width=0.99\columnwidth]{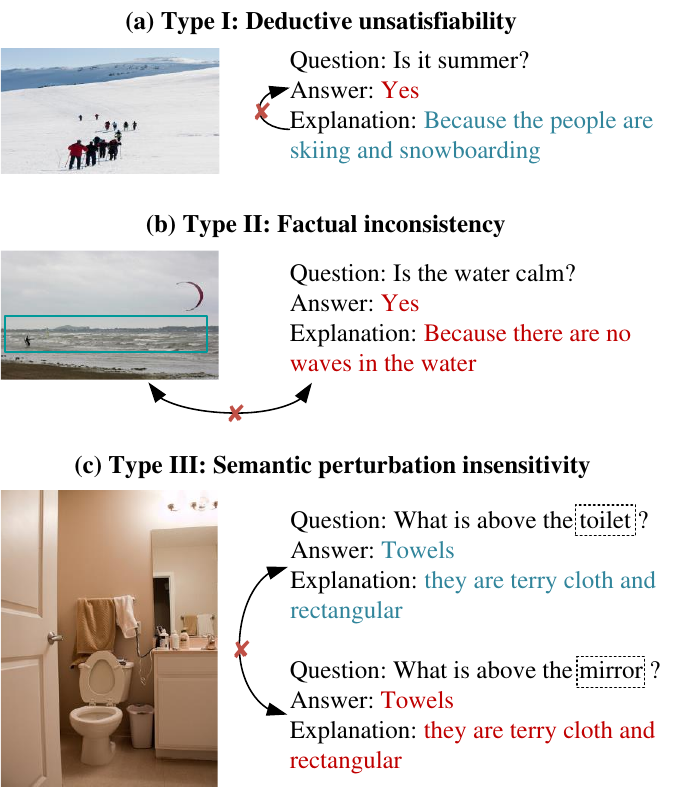} 
	\caption{Three types of logical errors in VQA-NLE: (a) The generated explanation does not logically lead to the answer; (b) The model falsifies its counterfactual explanation for the answer without considering the facts in image; and (c) The model infers the same explanation and answer for a statement and its opposite semantics.}
	\label{example}
\end{figure}

In contrast, natural language explanation (NLE) can provide the decision-making process of the model for users by generating a natural language sentence~\cite{camburu2018snli,park2018multimodal}, which is more accessible to understand. NLE can also improve the ability of large models to perform complex reasoning~\cite{wei2022chain}. Post-hoc NLE methods have achieved good performance on VQA~\cite{park2018multimodal,kayser2021vil,wu2019faithful}. They first gain answers for VQA by exploiting a vision-language (VL) model. Then the predicted answers along with the visual questions are fed into an explanation generation model to gain corresponding explanations. To reduce the high storage and memory requirements caused by the addition of the task model in post-hoc NLE methods, a self-rationalization method is proposed to predict an answer and explain it by formulating the answer prediction as a text generation task along with the explanation~\cite{sammani2022nlx}.

Despite their success, few VQA-NLE methods consider the faithfulness of the generated explanations and they suffer from the issue of logical inconsistency. As shown in Figure~\ref{example}, we manually inspected a large number of VQA-NLE samples with wrong answer predictions generated by the NLX-GPT model~\cite{sammani2022nlx} and found that: (a) the relationships between generated explanations and answers are deductive unsatisfiability (the generated explanation does not logically lead to the answer); (b) the explanations are inconsistent with the facts in corresponding images (the model falsifies its counterfactual explanations for the answer without considering the visual information); and (c) the generated explanations are insensitivity with semantic perturbations on visual questions (the model fails to recognize the semantic change caused by changing only several words or visual objects). These findings raise fundamental questions on the role of explanations in VQA: {\it How to improve the faithfulness of explanations and reduce inconsistencies between explanations and visual question answers? }

To address the above challenges, in this paper, we propose a \textbf{M}ulti-level \textbf{C}ontrastive \textbf{L}earning based natural language \textbf{E}xplanation (MCLE) framework which can learn discriminative representations from semantic-level, image-level, and instance-level factual and counterfactual samples. MCLE can encourage faithful explanations to be close to their corresponding visual questions and answers while to be far from other counterfactual (negative) samples. Specifically, MCLE consists of a vision-language (VL) model and a multi-level contrastive learning (CL) network. In our VL model, different from previous works~\cite{sammani2022nlx,suo2023s3c}, we consider the VQA-NLE task as a chain-of-thought (COT) generation task~\cite{wei2022chain}, where the answer is produced after the explanation. In our multi-level CL network, three core modules are designed to learn high-quality representations to guide the model to generate faithful explanation, i.e., the semantic-level CL (SemanticCL) for deductive satisfiability, the image-level CL (ImageCL) for factual consistency, and the instance-level CL (InstanceCL) for semantic perturbation sensitivity. Powered with the COT strategy and the multi-level CL, the MCLE effectively models the logical relationships and promotes the logical consistency between explanations and visual question answers. In terms of both automatic measures and human evaluation, our MCLE outperforms the state-of-the-art models for the VQA-NLE task on two widely used datasets, and improves the faithfulness of the generated explanations. 

In summary, we make the following contributions:
\begin{itemize}
	\item We propose a multi-level contrastive learning (MCLE) framework, i.e., semantic-level, image-level, and instance-level, to perform discriminative representation learning, which improves logical consistency and faithfulness over VQA-NLE generation task.
	\item We propose a chain-of-thought generation strategy in the vision-language model for VQA-NLE, which boosts the accuracy of predicted answers while improves the reliability of generated explanations.
	\item The proposed MCLE achieves new state-of-the-art performance on VQA-X and A-OKVQA benchmark datasets. Ablation analysis and case study are conducted to help understand the working of MCLE.
\end{itemize}

\section{Related Work}
\noindent
\textbf{Explainability in VQA} Given a question about an image, the goal of visual question answering is to generate an answer from both text and image information. It is firstly proposed by \cite{malinowski2014multi} and many approaches have been proposed such as joint embedding~\cite{dong2018predicting,yao2019hierarchy}, attention mechanisms~\cite{anderson2018bottom,lu2016hierarchical}, memory networks~\cite{ma2018visual,xiong2016dynamic} and graph neural networks~\cite{kipf2016semi,velickovic2017graph}. However, the reasoning process of the VQA models remains incomprehensible. Visualization technologies have been applied to achieve visual explanation~\cite{selvaraju2017grad,patro2019u}, but it has only limited expressiveness~\cite{wu2019self}. In contrast, text explanations formulated in \cite{park2018multimodal} are conducted on the VQA-NLE datasets and it utilizes human annotations to inspire the decision-making process of VQA models. \cite{kayser2021vil} combines a pre-trained language model and a VL model to generate free-text explanations while \cite{yang2022chunk} uses stronger VL models~\cite{li2020oscar} and generation models~\cite{radford2019language}. \cite{sammani2022nlx} proposes a unified model which can simultaneously predict answers and explanations based on a pre-trained caption model. Recently, \cite{suo2023s3c} introduces a self-criticism strategy to model the logical relationship between answers and reasons. However, they still suffer from logical errors in VQA-NLE including deductive unsatisfiability, factual inconsistency, and semantic insensitivity.

\noindent
\textbf{Contrastive Learning} The contrastive learning (CL) proposed in \cite{hadsell2006dimensionality} has been extensively researched and shown impressive results in extracting powerful representations. Typical contrastive learning methods aim to learn representations by contrasting positive and negative pairs. Many researchers have attempted to incorporate the CL into their models in an unsupervised learning manner and they achieved great success. \cite{dosovitskiy2014discriminative} uses unlabeled instances for contrastive representation in visual recognition. \cite{khosla2020supervised,tian2020makes} utilize labeled data, benefitting tasks like VQA~\cite{kim2021self,liang2020learning}, image caption~\cite{dai2017contrastive,li2020context}, and visual grounding~\cite{zhang2020counterfactual}. \cite{zhang2021multi} employs multi-level CL for visual commonsense reasoning, while \cite{liang2020learning} enhances VQA model generalization. Our approach adopts a multi-level contrastive architecture to improve the logical consistency and reliability over VQA-NLE explanation generation task.

\begin{figure*}[t]
	\centering
	\includegraphics[width=0.95\textwidth]{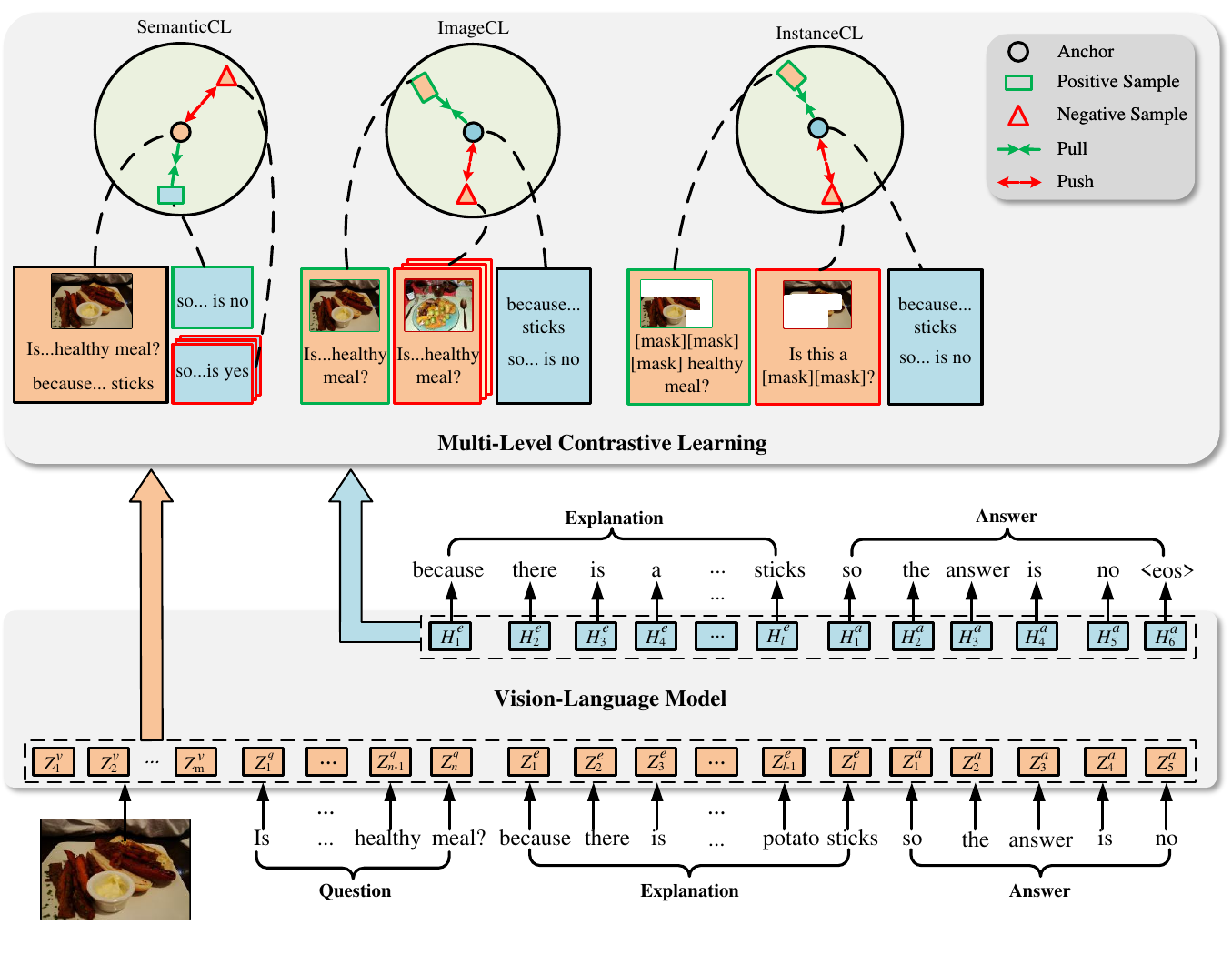} 
	\caption{The overall architecture of MCLE. It consists of a vision-language model with chain-of-thought generation strategy and a multi-level contrastive learning network (semantic-level, image-level, and instance-level).}
	\label{frame}
\end{figure*}

\section{Method}
In this section, we introduce our Multi-level Contrastive Learning based natural language Explanation (MCLE) framework. MCLE can improve the reliability of the rationales and strengthen the logical consistency between explanations and visual question answers. The overall architecture is shown in Figure~\ref{frame} where the MCLE comprises a vision-language model and a multi-level contrastive learning network with semantic-level, image-level, and instance-level. 

\subsection{Vision-Language Model}\label{vl}

\noindent
\textbf{Problem Formulation} Given an image $V$ and a natural language question $Q=(q_1,q_2,...,q_n)$, where $q_i$ represents the $i$-th word, the goal of VQA-NLE is to generate a corresponding free-text explanation with an answer. 

\noindent
\textbf{Text and Image Representation} Following previous works~\cite{sammani2022nlx}, we adopt the GPT-2~\cite{radford2019language} that pretrained on image caption task as our visual-language model and the CLIP~\cite{radford2021learning} as our image encoder. The question features $Z_Q=(Z^q_1,Z^q_2,...,Z^q_n)$ are obtained from the corresponding word embedding layer in GPT-2, where $Z^q_i\in \mathbb{R}^{d}$. The image features $Z_V=(Z^v_1,Z^v_2,...,Z^v_m)$ are encoded by CLIP, where $Z^v_i\in \mathbb{R}^{d}$.

\noindent
\textbf{Chain-of-thought Generation} To reduce the inconsistency between explanations and visual question answers, we introduce the chain-of-thought generation for VQA-NLE, which can mimic a rationale leading to the answer and provide an interpretable window into the decision-making progress. To inspire the model to generate faithful explanations and answers, the natural language `$because$' and `$so\;the\;answer\;is$' are as the prefixes of the ground-truth explanations and ground-truth answers, respectively. Then, like question features, the features of prefixed explanation and prefixed answer are obtained from the word embedding layer in GPT-2, where they are denoted by $Z_E=(Z^e_1,Z^e_2,...,Z^e_l)$ and $Z_A=(Z^a_1,Z^a_2,...,Z^a_5)$, respectively. By concatenating the prefixed explanation $Z_E$ with prefixed answer $Z_A$, we get the chain-of-thought features $Z_T=[Z_E;Z_A]$.  During training, all inputs (image $Z_V$, question $Z_Q$, chain-of-thought $Z_T$) are as a single sequence to the VL model. We train the VL model with the cross-entropy objective to generate the chain-of-thought sequence ${T} =  \{e_1,e_2,...,e_l,a_1,a_2,...a_5\}$ by minimizing:

\begin{equation}\label{COT_loss}
	\begin{split}
		\mathcal{L}_{vqa}&=- \log p \left({T} \mid Z_V,Z_Q \right)\\
		&= - \Big(\sum\nolimits_i \log p(e_{i} | H^e_i) + \sum\nolimits_i \log p(a_{i} | H^a_i) \Big)
	\end{split}
\end{equation} 

where 

\begin{equation*}\label{hidden_e}
	H^e_i= \text{VL}(Z_V,Z_Q,Z^e_1,Z^e_2,...,Z^e_{i-1};\theta)
\end{equation*} 
\begin{equation*}\label{hidden_a}
	H^a_i= \text{VL}(Z_V,Z_Q,Z_E,Z^a_1,Z^a_2,...,Z^a_{i-1};\theta)
\end{equation*} 
and $\theta$ denotes the parameters of the VL model. Note that, $H^e_1=\text{VL}(Z_V,Z_Q;\theta)$  and $H^a_1=\text{VL}(Z_V,Z_Q,Z_E;\theta)$.

Unlike other post-hoc methods, our explanation generation is solely based on visual questions and does not involve falsifying explanations based on the answer. Furthermore, the generated explanations can be used to prompt the generation of answers, which improves the logical consistency in VQA-NLE.

\subsection{Multi-Level Contrastive Learning Network}
Our multi-level contrastive learning network consists of three modules: SemanticCL, ImageCL, and InstanceCL. Following the contrastive learning framework in sequence to sequence learning~\cite{lee2020contrastive}, we maximize the similarity between the pair of anchor and positive (factual) sequence, while minimize the similarity between the pair of anchor and negative (counterfactual) as follows:

\begin{equation}
	\begin{split}
		\mathcal{L}_{CL}&=\text{CL}(\mathbf{x},\mathbf{S},\mathbf{y})\\
		&=- \log \frac{\exp \left(\text{sim}\left(\mathbf{e_x}, \mathbf{e_y}\right) / \tau\right)}{\sum_{\mathbf{\hat{x}}\in \mathbf{S}} \exp \left( \text{sim}\left(\mathbf{e_{\hat{x}}}, \mathbf{e_y}  \right) / \tau\right)}
	\end{split}
\end{equation}

where 

\begin{equation*} 
	\mathbf{e_x}=\xi\left(\mathbf{x};\theta \right)
\end{equation*} 
\begin{equation*} 
	\xi\left([x_1,...,x_t];\theta \right)=\text{AvgPool}([u_1,...,u_t])
\end{equation*} 
\begin{equation*} 
	u_t = \text{ReLU}(\mathbf{W}x_t+\mathbf{b})
\end{equation*} 

and $\mathbf{x}$ denotes positive (factual) sample, $\mathbf{S}$ denotes the set of negative (counterfactual) samples,  $\mathbf{y}$ denotes the anchor, and $\tau$ is the learned temperature parameter. The composition of affine transformation $\xi$ with the ReLU and AvgPool projects the sequences $[x_1,...,x_t]\in\mathbb{R}^{d \times t}$ onto the latent embedding space $\mathbf{e_x} \in \mathbb{R}^d$. The similarity $\text{sim}(\cdot,\cdot)$ measures between two sequences.

\noindent
\textbf{SemanticCL} To guide our VQA-NLE model to generate explanations that logically lead to the answers, we design a semantic-level CL module (SemanticCL) to learn the relationship between explanations and answers. Specifically, the ground-truth answer sequence is taken as the positive sample, the random sampled $K$ non-target answer sequences from the same batch are taken as the set of negative samples $\mathbf{S}$, and the combination of visual question and explanation is taken as the anchor. The contrastive loss in semanticCL is defined as follows:
\begin{equation}	
	\mathcal{L}_{semCL} = \text{CL}(\mathbf{x},\mathbf{S},\mathbf{y})\\
\end{equation}
where 
\begin{equation*}	
	\mathbf{x}={H_A}, \; \mathbf{S}=\{{\hat{H}_A}\}_{K}, \; \mathbf{y}={[Z_V;Z_Q;Z_E]}
\end{equation*}
and ${H_A}$, ${\hat{H}_A}$ denotes the positive and negative answer features obtained from the decoder hidden states of the VL model, respectively. ${Z_V,Z_Q,Z_E}$ denotes the image, question, and explanation features, respectively, which are obtained from the image encoder and word embedding layer of the VL model. $\{\cdot\}_K$ denotes the set of $K$ negative samples. Through the training, the corresponding explanation features are near to the ground-truth answer, while they are far away from the negative answers. In this way, the semanticCL helps learn the discriminative features between explanations and answers.

\noindent
\textbf{ImageCL} The image-level CL module (ImageCL) aims to guide the model to generate explanations closely related to the visual information, rather than to falsify counterfactual explanations according to the question only. Specifically, the combination of explanation and answer $[{H_E;H_A}]$ is taken as the anchor $\mathbf{y}$. The original image with question $[{Z_V;Z_Q}]$ is taken as the factual sample $\mathbf{x}$. The counterfactual image with question $[{\hat{Z}_V;Z_Q}]$ is taken as the counterfactual sample. 

During the counterfactual image sampling, we first calculate the score between original sample and other samples in the dataset by:
\begin{equation*}\label{score}
	score = \text{sim}\left(\mathbf{e_{\hat{q}}}, \mathbf{e_{q}}  \right) - \text{sim}\left(\mathbf{e_{\hat{a}}}, \mathbf{e_{a}}  \right)
\end{equation*}
where $\mathbf{e_{q}}$ ($\mathbf{e_{\hat{q}}}$) and $\mathbf{e_{a}}$ ($\mathbf{e_{\hat{a}}}$) are the latent embeddings of question and answer in original (counterfactual) sample, respectively. 

Then we select the images of $topK$ samples with the highest scores as the counterfactual images, which have a similar question but a different answer from the original sample, to guide the model to perceive the visual contents and eliminate the factual inconsistency caused by language bias. 

The contrastive loss in our ImageCL is defined as follows:
\begin{equation}	
	\mathcal{L}_{imgCL} = \text{CL}(\mathbf{x},\mathbf{S},\mathbf{y})\\
\end{equation}
where
\begin{equation*}	
	\mathbf{x}=[{Z_V;Z_Q}], \; \mathbf{S}=\{[{\hat{Z}_V;Z_Q}]\}_{K}, \; \mathbf{y}=[{H_E;H_A}]
\end{equation*}

Through the training, the explanations are near to the corresponding image, while they are far away from the counterfactual images. In this way, the ImageCL helps learn the discriminative features between explanations and images.

\noindent
\textbf{InstanceCL} To help VQA-NLE model perceive the semantic changes caused by fine-grained visual or text perturbations, we design an instance-level CL module (InstanceCL). In this module, a gradient-based counterfactual transformation strategy is adopted to synthesize factual and counterfactual samples. We apply the modified Grad-CAM~\cite{selvaraju2017grad} to derive the contribution of the $i$-th object and the $j$-th word to answer by the following functions:
\begin{equation}
	s(a,Z^v_i) = S(P_{vqa}(a),Z^v_i)=(\nabla_{Z^v_i}P_{vqa}(a))^\top \mathbf{1}
\end{equation}
\begin{equation}
	s(a,Z^q_j) = S(P_{vqa}(a),Z^q_j)=(\nabla_{Z^q_j}P_{vqa}(a))^\top \mathbf{1}
\end{equation}
\noindent
where $P_{vqa}(a)$ is the predicted answer probability of the ground truth answer, $Z^v_i$ is the $i$-th object features, $Z^q_j$ is the $j$-th word features, and $\mathbf{1}$ is an all ones vector. 

Obviously, if the score $s(a, \cdot)$ is higher, the contribution of the object $Z^v_i$ (or $Z^q_j$) to the answer is larger. Based on such contribution scores, the $topK$ objects and words with the highest scores are collected as the factual samples $[{Z^+_V;Z^+_Q}]$ while the counterfactual samples $[{Z^-_V;Z^-_Q}]$ are generated by masking the corresponding factual samples. The contrastive loss in InstanceCL is defined as follows:
\begin{equation}	
	\mathcal{L}_{insCL} = \text{CL}(\mathbf{x},\mathbf{S},\mathbf{y})\\
\end{equation}
where
\begin{equation*}	
	\mathbf{x}=[{Z^+_V;Z^+_Q}], \; \mathbf{S}=[{Z^-_V;Z^-_Q}], \; \mathbf{y}=[{H_E;H_A}]
\end{equation*}
and the union set $Z_V = Z^+_V \cup Z^-_V$, and  $Z_Q = Z^+_Q \cup Z^-_Q$. During training, explanations are near to the corresponding objects and words, while they are far away from unrelated objects and words. In this way, the ImageCL helps perceive the key fine-grained content in the image and question.

\subsection{Overall Loss}
The overall loss of our MCLE is:

\begin{equation}	
	\mathcal{L} =\mathcal{L}_{vqa}+	\alpha \mathcal{L}_{semCL}+\beta \mathcal{L}_{imgCL}+\gamma	\mathcal{L}_{insCL}\\
\end{equation}
\noindent
where $\alpha$, $\beta$, and $\gamma$ are the trade-off parameters. MCLE improves the logical consistency and reliability of the VQA-NLE task by jointly optimizing the main loss (the vision-language model) and three auxiliary losses (the multi-level contrastive learning network).

\section{Experiment}
\begin{table*}[t]
	\centering
	\resizebox{0.85\linewidth}{!}{
		\begin{tabular}{c|ccccccc|ccccccc}
			\toprule
			& \multicolumn{7}{c|}{VQA-X}                       & \multicolumn{7}{c}{A-OKVQA}                       \\ \midrule
			Approach   & B4   & M    & R    & C     & S    & Acc  & Human & B4   & M    & R    & C    & S    & Acc  & Human \\ \midrule
			PJ-X       & 19.5 & 18.2 & 43.4 & 71.3  & 15.1 & 76.4 & 65.4  & - & - & - & - & -  & -    & -  \\
			FME        & 24.4 & 19.5 & 47.4 & 88.8  & 17.9 & 75.5 & -     & -    & -    & -    & -    & -    &      & -     \\
			e-UC       & -    & -    & -    & -     & -    & -    & -     & 15.1 & 18.1 & 42.4 & 51.5 & 14.9 & 25.6 & 44.1  \\
			NLX-GPT    & 25.6 & 21.5 & 48.7 & 97.2  & 20.2 & 83.1 & 70.2  & 20.1 & 17.0 & 46.3 & 65.4 & 15.8 & 28.7 & 46.9  \\
			$\text{S}^3\text{C}$        & 27.8 & 22.8 & 50.7 & 104.4 & 21.5 & 85.6 & 77.4  & 22.5 & 18.5 & 48.4 & 74.4 & 18.1 & 33.5 & 54.7  \\ \midrule
			MCLE (ours) & \textbf{28.6} & \textbf{24.2} & \textbf{52.3} & \textbf{106.7} & \textbf{23.9} & \textbf{87.7} & \textbf{80.8}  & \textbf{23.1} & \textbf{19.7} & \textbf{50.1} & \textbf{76.6} & \textbf{19.7} & \textbf{34.7} &\textbf{57.9}   \\ \bottomrule
		\end{tabular}
	}
	\caption{Comparison with the state-of-the-art methods on the VQA-X and A-OKVQA datasets in the scenario of ``unfiltered'' scores. (``unfiltered'' indicates that the explanations are evaluated regardless of whether the answer is true or false, while ``filtered'' is to only consider the explanations that have correct answers.) The B4, M, R, C, S, Acc, and Human are short for BLEU-4, METEOR, ROUGE-L, CIDEr, SPICE, Answer Accuracy, and Human Evaluation, respectively.}
	\label{tb:unfiltered}
\end{table*}

\subsection{Datasets}
Following~\cite{suo2023s3c}, we conduct empirical experiments on two widely used VQA-NLE benchmarks. 

\noindent
\textbf{VQA-X}~\cite{park2018multimodal} is collected from the VQA dataset~\cite{antol2015vqa} and provides additional explanations for the answers. It consists of 28K images and 33K QA pairs, split into 29K/1.4K/1.9K for training, validation, and testing, respectively. 

\noindent
\textbf{A-OKVQA}~\cite{schwenk2022okvqa} is collected from the COCO dataset~\cite{lin2014microsoft}. It includes about 25K Question/Answer/Rationale triplets, split into 17.1K/1.1K/6.7K for training, validation, and testing, respectively.

\subsection{Evaluation Measures}
\textbf{Automatic Evaluation} Following~\cite{suo2023s3c}, the generated explanations are evaluated in terms of the metrics BLUE~\cite{papineni2002bleu}, METEOR~\cite{denkowski2014meteor}, ROUGE-L~\cite{lin2004rouge}, SPICE~\cite{anderson2016spice}, and CIDEr~\cite{vedantam2015cider}. The predicted answers are evaluated in terms of the metric Accuracy. 

\noindent
\textbf{Human Evaluation} Following~\cite{suo2023s3c}, we use human evaluations to measure the faithfulness and logicality of the explanations, since they are not always reflected by the automatic metrics~\cite{kayser2021vil}. Specifically, three human evaluators are employed to determine each generated explanation whether it is consistent with the answer and they select an option from [yes, weak yes, weak no, no], corresponding to scores [1, 2/3, 1/3, 0], respectively. We compute an average among total scores of all test samples to obtain the final human evaluation score. Meanwhile, these evaluators are asked to choose the reason for unqualified explanations: deductive unsatisfiability, factual inconsistency, and semantic perturbation insensitivity (see Figure~\ref{example}).

\subsection{Experimental Setup}

\noindent
\textbf{Baselines} We compare with five strong baselines. 

\begin{itemize}
  \item \textbf{PJ-X}~\cite{park2018multimodal} is a post-hoc method by creating attention features to guide the generation of textual explanations.
  \item \textbf{FME}~\cite{wu2019faithful} uses the Grad-CAM~\cite{selvaraju2017grad} to generate the explanations which can be traced back to the relevant object set. 
  \item \textbf{e-UG}~\cite{kayser2021vil} generates the explanations by combining UNITER~\cite{chen2020uniter} and GPT-2. 
  \item \textbf{NLX-GPT}~\cite{sammani2022nlx} can simultaneously predict an answer and explain it by formulating the answer prediction as a text generation task along with the explanation.
  \item \textbf{$\text{S}^3\text{C}$}~\cite{suo2023s3c} is a self-critical VQA-NLE method that can model the logical relationships between answer-explanation pairs.
\end{itemize}

\noindent
\textbf{Implementation} We conduct all experiments on NVIDIA GTX 3080 Ti GPUs with PyTorch 1.9.0. We take the GPT-2 model that pre-trained on image caption task~\cite{sammani2022nlx} as our vision-language model backbone. The temperature $\tau$ is set to 0.2. The hyperparameters top-$K$ in the multi-level CL (SemanticCL, ImageCL, and InstanceCL) are set to (3, 3, 2), respectively. The trade-off parameters $\alpha$, $\beta$, and $\gamma$ are set to 0.1, 0.2, and 0.2, respectively. The maximum length of text sentence cuts at 40, batch size is 16, and training epoch is 30. See our released code at  \url{https://github.com/laichengen/MCLE}

\begin{table}[t]
	\centering
	\resizebox{\linewidth}{!}{
		\begin{tabular}{c|ccccccc}
			\toprule
			& \multicolumn{7}{c}{{VQA-X}} \\ \midrule
			Approach & B4 & M & R & C & S & Acc & Human \\ \midrule
			PJ-X & 22.7 & 19.7 & 46.0 & 82.7  &  17.1 & 76.4 & 69.3 \\
			FME &  23.1 & 20.4 & 47.1 &87.0  & 18.4 & 75.5 & - \\
			e-UG & 23.2 & 22.1 &  45.7 & 74.1   &  20.1 & 80.5 & 71.4 \\
			NLX-GPT & 28.5 & 23.1 & 51.5 &  110.6  &22.1   & 83.1 & 73.7 \\ 
			$\text{S}^3\text{C}$ & {30.7} & {23.9} & {52.1} & {116.7} & {23.0} & {85.6} & {79.2} \\ 
			\midrule
			MCLE (ours) & \textbf{31.2} & \textbf{24.2} & \textbf{53.1} & \textbf{118.3} & \textbf{24.2} & \textbf{87.7} & \textbf{81.3} \\
			\bottomrule
		\end{tabular}
	}
	\caption{Comparison with the state-of-the-art methods on the VQA-X dataset in the scenario of ``filtered'' scores. (``unfiltered'' indicates that the explanations are evaluated regardless of whether the answer is true or false, while ``filtered'' is to only consider the explanations that have correct answers.) }
	\label{tb:filtered}
\end{table}

\subsection{Main Results on Automatic Evaluation}

\noindent
\textbf{Unfiltered Scenario} The performance comparison of different methods is shown in Table~\ref{tb:unfiltered}, where the best results are in boldface. We have the following observations. We observe that our MCLE achieves the new state-of-the-art performance on two VQA-NLE datasets (VQA-X and A-OKVQA). Specifically, our MCLE outperforms the best baseline with 2.4\% improvement in terms of SPICE on the VQA-X dataset, while with 2.2\% improvement in terms of CIDEr on the A-OKVQA. Furthermore, our MCLE gets the best result over all of the 12 automatic evaluation settings on the two datasets, with average 1.7\% and 1.4\% improvements respectively. This shows that our model can generate more reliable explanations. As for the accuracy of answers (see the column of ``Acc''), our MCLE can simultaneously boost the precision of answers and corresponding explanations. These improvements of our MCLE over baselines could be attributed to two reasons: i) MCLE adopts an explain-then-predict framework with chain-of-thought generation strategy, which can mimic a rationale leading to the answer; and ii) The multi-level contrastive learning network is able to learn high-quality representations to guide the model to generate logical consistency explanations.  

\noindent
\textbf{Filtered Scenario} To verify the algorithm's validity, we follow~\cite{kayser2021vil} to report the filtered scores on the VQA-X dataset as shown in Table~\ref{tb:filtered}. Our MCLE achieves a new state-of-the-art on VQA-X. Specifically, our method can outperform the baseline methods with 1.6\% improvement in terms of SPICE and with 1.4\% improvement in terms of answer accuracy.

\subsection{Main Results on Human Evaluation}

\noindent
\textbf{Unfiltered and Filtered Scenarios} We conduct the human evaluation to evaluate the correctness and faithfulness of generated explanations. As shown in Table~\ref{tb:unfiltered}, compared to other five methods, the Human score of our MCLE is improved by 3.4 points on VQA-X in the scenario of unfiltered scores. As shown in Table~\ref{tb:filtered}, compared to other five methods, the Human score of our MCLE is improved by 2.1 points on VQA-X in the scenario of filtered scores. 

\noindent
\textbf{Logical Errors} Moreover, we also ask the human evaluators to select the reasons for each unqualified explanation on the VQA-X dataset. As shown in Table~\ref{tb:shortcoming}, the insufficient explanations caused by deductive unsatisfiability are reduced by 1.8\%, the irrelevant explanations caused by factual inconsistency are reduced by 1.6\%, and the meaningless explanations caused by semantic perturbation insensitivity are reduced by 0.4\%. These results indicate that our MCLE can obtain relatively better rationales and empirically confirm the effectiveness of our method.

\begin{table}[t]
	\centering
	\resizebox{0.76\linewidth}{!}{
		\begin{tabular}{c|ccc}
			\toprule
               & \multicolumn{3}{c}{VQA-X} \\ \midrule
			Approach &    Type I    &   Type II     &    Type III   \\ \midrule
			PJ-X                      &  28.4\%          &  21.1\%       & 9.2\%                      \\
			e-UG                      &   25.4\%          &  22.8\%      & 8.7\%                      \\
			NLX-GPT                   &  22.2\%         &    20.3\%      & 9.1\%                      \\
			$\text{S}^3\text{C}$      &  18.9\%         &  17.3\%        & 8.2\%                      \\ \midrule
			MCLE (ours)               & \textbf{17.1\% }          & \textbf{15.7\% }       & \textbf{7.8\%  }                    \\ \bottomrule
		\end{tabular}
	}
	\caption{The main reason of unqualified explanations on the VQA-X dataset. Three types of logical errors: (a) Type I: Deductive unsatisfiability, (b) Type II: Factual inconsistency, and (c) Type III: Semantic perturbation insensitivity (see Figure~\ref{example}).}
	\label{tb:shortcoming}
\end{table}

\begin{table*}[t]
	\centering
	\resizebox{0.86\linewidth}{!}{
		\begin{tabular}{l|ccccccc|ccccccc}
			\toprule
			& \multicolumn{7}{c|}{VQA-X}                       & \multicolumn{7}{c}{A-OKVQA}                       \\ \midrule
			Approach   & B4   & M    & R    & C     & S    & Acc  & Human & B4   & M    & R    & C    & S    & Acc  & Human \\ \midrule
			MCLE       & \textbf{28.6} & \textbf{24.2} & \textbf{52.3} & \textbf{106.7} & \textbf{23.9} & \textbf{87.7} & \textbf{80.8}  & \textbf{23.1} & \textbf{19.7} & \textbf{50.1} & \textbf{76.6} & \textbf{19.7} & \textbf{34.7} &\textbf{57.9}  \\ \midrule
			w/o  COT       & 27.8 & 23.6 & 51.7 & 105.5 & 23.1 & 86.5 & 79.4  & 22.5 & 18.8 & 48.6 & 75.7 & 19.1 & 33.7 & 56.2  \\ 
			w/o SemanticCL        & 27.1 & 23.5 & 51.1 & 104.6  & 22.3 & 86.1 & 78.8     & 22.3    & 18.2    & 48.6    & 75.9    & 18.8    &  33.1    & 55.8     \\
			w/o ImageCL      & 26.9 & 21.5 & 48.7 & 103.3  & 21.7 & 85.8 & 77.6  & 20.3 & 17.8 & 47.8 & 75.4 & 16.8 & 31.7 & 53.9  \\
			w/o InstanceCL     & 27.3    & 22.8   & 49.9    & 105.1     & 22.8    & 86.4    & 79.2     & 21.8 & 18.1 & 49.4 & 75.5 & 18.3 & 32.9 & 54.6  \\ 
			
			w/o All & 25.9 & 22.1 & 48.3 & 98.1 & 20.9 & 83.5 & 71.8  & 20.1 & 16.8 & 46.2 & 66.5 & 16.2 & 28.7 & 47.2  \\ \bottomrule
		\end{tabular}
	}
	\caption{Ablated results of our MCLE and its key components, chain-of-thought (COT) generation strategy and the multi-level contrastive learning network (semantic-level, image-level, and instance-level).}
	\label{tb:ablation}
\end{table*}

\begin{figure*}[t]
	\centering
	\includegraphics[width=0.98\textwidth]{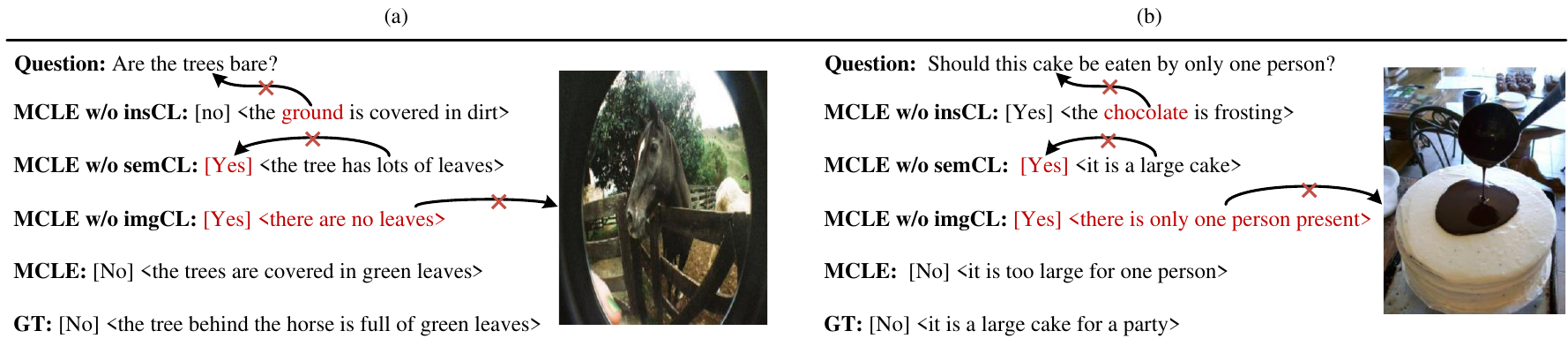} 
	\caption{Case study on the generated explanations on the VQA-X dataset. The $[\cdot]$ and $<\cdot>$ indicate answers and explanations respectively. We show the results of our full MCLE model and its three variants. GT denotes by the ground truth.}
	\label{case}
\end{figure*}

\subsection{Ablation Studies}
The ablation results of the full MCLE model and its five variants are shown in Table~\ref{tb:ablation}. From the results, we have the following findings. 

Firstly, for the effectiveness of the chain-of-thought (COT) generation strategy which mimics a rationale leading to the answer and provides an interpretable window into the decision-making progress of the model, MCLE w/o CTG performs worse than MCLE. For example, CIDEr reduces by 1.2\% on VQA-X. It verifies that the COT strategy is important for the model to improve the explanation's logical consistency.

Secondly, for the effectiveness of the SemanticCL strategy which learns the discriminative features between explanations and answers, MCLE w/o SemanticCL performs worse than MCLE. For example, SPICE reduces by 1.6\% on VQA-X. It verifies that the SemanticCL is important to guide the model to generate explanations that logically lead to the answers.

Thirdly, for the effectiveness of the ImageCL strategy which learns the discriminative features between explanations and images, MCLE w/o ImageCL performs worse than MCLE. For example, ROUGE-L reduces by 3.6\% on VQA-X. It verifies that the ImageCL is important to guide the model to generate explanations closely related to the visual information, rather than falsifying counterfactual explanations caused by language bias.

Fourthly, for the effectiveness of the InstanceCL strategy which perceives the semantic changes caused by fine-grained visual and text perturbation, MCLE w/o InstanceCL performs worse than MCLE. For example, ROUGE-L reduces by 2.4\% on VQA-X. It verifies that the InstanceCL is important to guide the model to perceive the key fine-grained content in image and question.

Obviously, MCLE w/o All is the worst. For example, CIDEr reduces by 8.6\% on VQA-X. It further shows that both of the chain-of-thought (COT) generation strategy and multi-level contrastive learning network in our MCLE contribute to the performance improvements.

\subsection{Case Studies}
We have quantitatively demonstrated the effectiveness of our MCLE by comparing with five state-of-the-art methods, and conducted detailed ablation study on the contributions from the core components of chain-of-thought generation strategy and multi-level contrastive learning network. In this section, to obtain an intuitive understanding of how the proposed MCLE works, we show typically qualitative results from the NLX-GPT module~\cite{sammani2022nlx} on the VQA-X dataset by comparing the generated explanations and answers of MCLE and its variants.

As shown in Figure~\ref{case}(a), the MCLE w/o InstanceCL generates an explanation that is relevant to the word ``ground'' rather than ``trees'', maybe caused by the language bias in the dataset (e.g., ``Are the ground bare?''). For the MCLE w/o SemanticCL, although the generated explanation correctly identifies the tree having lots of leaves, the predicted answer is wrongly ``yes'' (answering the trees as bare for the question), which is contradictory to the explanation and suffers from deductive unsatisfiability (Type I logical error in Figure~\ref{example}). The MCLE w/o ImageCL falsifies the explanation that the tree in the image has no leaves, which is inconsistent with the factual image and suffers from factual inconsistency  (Type II logical error). For our full MCLE model with all three-level contrastive learning (CL) components, it correctly generates more faithful rationales and predicts logically consistent answers. This shows that the multi-level CL can help vision-language models improve the logical consistency between explanations and visual question answers. Figure~\ref{case}(b) is another similar case to help understand the working of our proposed MCLE framework and the contributions from the three-level CL network.


\subsection{Conclusion}
We proposed a novel self-supervised {M}ulti-level {C}ontrastive {L}earning based natural language {E}xplanation model (MCLE) for VQA with semantic-level, image-level, and instance-level factual and counterfactual (negative) samples. MCLE can learn discriminative features and align the feature spaces from explanations with visual questions and answers to generate more consistent explanations. Our model improves consistent and faithful explanations while reduces the deductive unsatisfiability, factual inconsistency, and semantic perturbation insensitivity. From both automatic measures and human evaluations, our MCLE achieves a new state-of-the-art on VQA-NLE task. 

\subsection{Acknowledgment}
We are grateful to the anonymous reviewers, SPC, AC, and PC chairs for their great efforts. This work is funded by the NSFC 62306220.

\bibliography{aaai24}

\begin{thebibliography}{44}
\providecommand{\natexlab}[1]{#1}

\bibitem[{Anderson et~al.(2016)Anderson, Fernando, Johnson, and
  Gould}]{anderson2016spice}
Anderson, P.; Fernando, B.; Johnson, M.; and Gould, S. 2016.
\newblock SPICE: semantic propositional image caption evaluation.
\newblock In \emph{European Conference on Computer Vision}, 382--398. Springer,
  Springer Nature.

\bibitem[{Anderson et~al.(2018)Anderson, He, Buehler, Teney, Johnson, Gould,
  and Zhang}]{anderson2018bottom}
Anderson, P.; He, X.; Buehler, C.; Teney, D.; Johnson, M.; Gould, S.; and
  Zhang, L. 2018.
\newblock Bottom-up and top-down attention for image captioning and visual
  question answering.
\newblock In \emph{Proceedings of the IEEE conference on computer vision and
  pattern recognition}, 6077--6086.

\bibitem[{Antol et~al.(2015)Antol, Agrawal, Lu, Mitchell, Batra, Zitnick, and
  Parikh}]{antol2015vqa}
Antol, S.; Agrawal, A.; Lu, J.; Mitchell, M.; Batra, D.; Zitnick, C.~L.; and
  Parikh, D. 2015.
\newblock Vqa: Visual question answering.
\newblock In \emph{Proceedings of the IEEE international conference on computer
  vision}, 2425--2433.

\bibitem[{Camburu et~al.(2018)Camburu, Rockt{\"a}schel, Lukasiewicz, and
  Blunsom}]{camburu2018snli}
Camburu, O.-M.; Rockt{\"a}schel, T.; Lukasiewicz, T.; and Blunsom, P. 2018.
\newblock e-snli: Natural language inference with natural language
  explanations.
\newblock \emph{Advances in Neural Information Processing Systems}, 31.

\bibitem[{Chen and Zhao(2022)}]{chen2022rex}
Chen, S.; and Zhao, Q. 2022.
\newblock Rex: Reasoning-aware and grounded explanation.
\newblock In \emph{Proceedings of the IEEE/CVF Conference on Computer Vision
  and Pattern Recognition}, 15586--15595.

\bibitem[{Chen et~al.(2020)Chen, Li, Yu, El~Kholy, Ahmed, Gan, Cheng, and
  Liu}]{chen2020uniter}
Chen, Y.-C.; Li, L.; Yu, L.; El~Kholy, A.; Ahmed, F.; Gan, Z.; Cheng, Y.; and
  Liu, J. 2020.
\newblock Uniter: Universal image-text representation learning.
\newblock In \emph{European conference on computer vision}, 104--120. Springer.

\bibitem[{Dai and Lin(2017)}]{dai2017contrastive}
Dai, B.; and Lin, D. 2017.
\newblock Contrastive learning for image captioning.
\newblock \emph{Advances in Neural Information Processing Systems}, 30.

\bibitem[{Denkowski and Lavie(2014)}]{denkowski2014meteor}
Denkowski, M.; and Lavie, A. 2014.
\newblock Meteor universal: Language specific translation evaluation for any
  target language.
\newblock In \emph{Proceedings of the ninth workshop on statistical machine
  translation}, 376--380.

\bibitem[{Dong, Li, and Snoek(2018)}]{dong2018predicting}
Dong, J.; Li, X.; and Snoek, C.~G. 2018.
\newblock Predicting visual features from text for image and video caption
  retrieval.
\newblock \emph{IEEE Transactions on Multimedia}, 20(12): 3377--3388.

\bibitem[{Dosovitskiy et~al.(2014)Dosovitskiy, Springenberg, Riedmiller, and
  Brox}]{dosovitskiy2014discriminative}
Dosovitskiy, A.; Springenberg, J.~T.; Riedmiller, M.; and Brox, T. 2014.
\newblock Discriminative unsupervised feature learning with convolutional
  neural networks.
\newblock \emph{Advances in neural information processing systems}, 27.

\bibitem[{Hadsell, Chopra, and LeCun(2006)}]{hadsell2006dimensionality}
Hadsell, R.; Chopra, S.; and LeCun, Y. 2006.
\newblock Dimensionality reduction by learning an invariant mapping.
\newblock In \emph{2006 IEEE computer society conference on computer vision and
  pattern recognition (CVPR'06)}, volume~2, 1735--1742. IEEE.

\bibitem[{Kayser et~al.(2021)Kayser, Camburu, Salewski, Emde, Do, Akata, and
  Lukasiewicz}]{kayser2021vil}
Kayser, M.; Camburu, O.-M.; Salewski, L.; Emde, C.; Do, V.; Akata, Z.; and
  Lukasiewicz, T. 2021.
\newblock e-vil: A dataset and benchmark for natural language explanations in
  vision-language tasks.
\newblock In \emph{Proceedings of the IEEE/CVF international conference on
  computer vision}, 1244--1254.

\bibitem[{Khosla et~al.(2020)Khosla, Teterwak, Wang, Sarna, Tian, Isola,
  Maschinot, Liu, and Krishnan}]{khosla2020supervised}
Khosla, P.; Teterwak, P.; Wang, C.; Sarna, A.; Tian, Y.; Isola, P.; Maschinot,
  A.; Liu, C.; and Krishnan, D. 2020.
\newblock Supervised contrastive learning.
\newblock \emph{Advances in neural information processing systems}, 33:
  18661--18673.

\bibitem[{Kim et~al.(2021)Kim, Jeong, Kim, Kang, and Kwak}]{kim2021self}
Kim, S.; Jeong, S.; Kim, E.; Kang, I.; and Kwak, N. 2021.
\newblock Self-supervised pre-training and contrastive representation learning
  for multiple-choice video qa.
\newblock In \emph{Proceedings of the AAAI Conference on Artificial
  Intelligence}, volume~35, 13171--13179.

\bibitem[{Kipf and Welling(2016)}]{kipf2016semi}
Kipf, T.~N.; and Welling, M. 2016.
\newblock Semi-supervised classification with graph convolutional networks.
\newblock \emph{arXiv preprint arXiv:1609.02907}.

\bibitem[{Lee, Lee, and Hwang(2020)}]{lee2020contrastive}
Lee, S.; Lee, D.~B.; and Hwang, S.~J. 2020.
\newblock Contrastive Learning with Adversarial Perturbations for Conditional
  Text Generation.
\newblock In \emph{International Conference on Learning Representations}.

\bibitem[{Li et~al.(2020{\natexlab{a}})Li, Yin, Li, Zhang, Hu, Zhang, Wang, Hu,
  Dong, Wei et~al.}]{li2020oscar}
Li, X.; Yin, X.; Li, C.; Zhang, P.; Hu, X.; Zhang, L.; Wang, L.; Hu, H.; Dong,
  L.; Wei, F.; et~al. 2020{\natexlab{a}}.
\newblock Oscar: Object-semantics aligned pre-training for vision-language
  tasks.
\newblock In \emph{Computer Vision--ECCV 2020: 16th European Conference,
  Glasgow, UK, August 23--28, 2020, Proceedings, Part XXX 16}, 121--137.
  Springer.

\bibitem[{Li et~al.(2020{\natexlab{b}})Li, Tran, Mai, Lin, and
  Yuille}]{li2020context}
Li, Z.; Tran, Q.; Mai, L.; Lin, Z.; and Yuille, A.~L. 2020{\natexlab{b}}.
\newblock Context-aware group captioning via self-attention and contrastive
  features.
\newblock In \emph{Proceedings of the IEEE/CVF conference on computer vision
  and pattern recognition}, 3440--3450.

\bibitem[{Liang et~al.(2020)Liang, Jiang, Hu, and Zhu}]{liang2020learning}
Liang, Z.; Jiang, W.; Hu, H.; and Zhu, J. 2020.
\newblock Learning to contrast the counterfactual samples for robust visual
  question answering.
\newblock In \emph{Proceedings of the 2020 conference on empirical methods in
  natural language processing (EMNLP)}, 3285--3292.

\bibitem[{Lin(2004)}]{lin2004rouge}
Lin, C.-Y. 2004.
\newblock Rouge: A package for automatic evaluation of summaries.
\newblock In \emph{Text summarization branches out}, 74--81.

\bibitem[{Lin et~al.(2014)Lin, Maire, Belongie, Hays, Perona, Ramanan,
  Doll{\'a}r, and Zitnick}]{lin2014microsoft}
Lin, T.-Y.; Maire, M.; Belongie, S.; Hays, J.; Perona, P.; Ramanan, D.;
  Doll{\'a}r, P.; and Zitnick, C.~L. 2014.
\newblock Microsoft COCO: Common Objects in Context.
\newblock In \emph{European Conference on Computer Vision}, 740--755. Springer.

\bibitem[{Lu et~al.(2016)Lu, Yang, Batra, and Parikh}]{lu2016hierarchical}
Lu, J.; Yang, J.; Batra, D.; and Parikh, D. 2016.
\newblock Hierarchical question-image co-attention for visual question
  answering.
\newblock \emph{Advances in neural information processing systems}, 29.

\bibitem[{Ma et~al.(2018)Ma, Shen, Dick, Wu, Wang, Van~den Hengel, and
  Reid}]{ma2018visual}
Ma, C.; Shen, C.; Dick, A.; Wu, Q.; Wang, P.; Van~den Hengel, A.; and Reid, I.
  2018.
\newblock Visual question answering with memory-augmented networks.
\newblock In \emph{Proceedings of the IEEE conference on computer vision and
  pattern recognition}, 6975--6984.

\bibitem[{Malinowski and Fritz(2014)}]{malinowski2014multi}
Malinowski, M.; and Fritz, M. 2014.
\newblock A multi-world approach to question answering about real-world scenes
  based on uncertain input.
\newblock \emph{Advances in neural information processing systems}, 27.

\bibitem[{Papineni et~al.(2002)Papineni, Roukos, Ward, and
  Zhu}]{papineni2002bleu}
Papineni, K.; Roukos, S.; Ward, T.; and Zhu, W.-J. 2002.
\newblock Bleu: a method for automatic evaluation of machine translation.
\newblock In \emph{Proceedings of the 40th annual meeting of the Association
  for Computational Linguistics}, 311--318.

\bibitem[{Park et~al.(2018)Park, Hendricks, Akata, Rohrbach, Schiele, Darrell,
  and Rohrbach}]{park2018multimodal}
Park, D.~H.; Hendricks, L.~A.; Akata, Z.; Rohrbach, A.; Schiele, B.; Darrell,
  T.; and Rohrbach, M. 2018.
\newblock Multimodal explanations: Justifying decisions and pointing to the
  evidence.
\newblock In \emph{Proceedings of the IEEE conference on computer vision and
  pattern recognition}, 8779--8788.

\bibitem[{Patro et~al.(2019)Patro, Lunayach, Patel, and
  Namboodiri}]{patro2019u}
Patro, B.~N.; Lunayach, M.; Patel, S.; and Namboodiri, V.~P. 2019.
\newblock U-cam: Visual explanation using uncertainty based class activation
  maps.
\newblock In \emph{Proceedings of the IEEE/CVF International Conference on
  Computer Vision}, 7444--7453.

\bibitem[{Radford et~al.(2021)Radford, Kim, Hallacy, Ramesh, Goh, Agarwal,
  Sastry, Askell, Mishkin, Clark et~al.}]{radford2021learning}
Radford, A.; Kim, J.~W.; Hallacy, C.; Ramesh, A.; Goh, G.; Agarwal, S.; Sastry,
  G.; Askell, A.; Mishkin, P.; Clark, J.; et~al. 2021.
\newblock Learning transferable visual models from natural language
  supervision.
\newblock In \emph{International conference on machine learning}, 8748--8763.
  PMLR.

\bibitem[{Radford et~al.(2019)Radford, Wu, Child, Luan, Amodei, Sutskever
  et~al.}]{radford2019language}
Radford, A.; Wu, J.; Child, R.; Luan, D.; Amodei, D.; Sutskever, I.; et~al.
  2019.
\newblock Language models are unsupervised multitask learners.
\newblock \emph{OpenAI blog}, 1(8): 9.

\bibitem[{Sammani, Mukherjee, and Deligiannis(2022)}]{sammani2022nlx}
Sammani, F.; Mukherjee, T.; and Deligiannis, N. 2022.
\newblock NLX-GPT: A Model for Natural Language Explanations in Vision and
  Vision-Language Tasks.
\newblock In \emph{2022 IEEE/CVF Conference on Computer Vision and Pattern
  Recognition (CVPR)}, 8312--8322. IEEE.

\bibitem[{Schwenk et~al.(2022)Schwenk, Khandelwal, Clark, Marino, and
  Mottaghi}]{schwenk2022okvqa}
Schwenk, D.; Khandelwal, A.; Clark, C.; Marino, K.; and Mottaghi, R. 2022.
\newblock A-okvqa: A benchmark for visual question answering using world
  knowledge.
\newblock In \emph{European Conference on Computer Vision}, 146--162. Springer.

\bibitem[{Selvaraju et~al.(2017)Selvaraju, Cogswell, Das, Vedantam, Parikh, and
  Batra}]{selvaraju2017grad}
Selvaraju, R.~R.; Cogswell, M.; Das, A.; Vedantam, R.; Parikh, D.; and Batra,
  D. 2017.
\newblock Grad-cam: Visual explanations from deep networks via gradient-based
  localization.
\newblock In \emph{Proceedings of the IEEE international conference on computer
  vision}, 618--626.

\bibitem[{Suo et~al.(2023)Suo, Sun, Liu, Gao, Wang, Zhang, and Wu}]{suo2023s3c}
Suo, W.; Sun, M.; Liu, W.; Gao, Y.; Wang, P.; Zhang, Y.; and Wu, Q. 2023.
\newblock S3C: Semi-Supervised VQA Natural Language Explanation via
  Self-Critical Learning.
\newblock In \emph{Proceedings of the IEEE/CVF Conference on Computer Vision
  and Pattern Recognition}, 2646--2656.

\bibitem[{Tian et~al.(2020)Tian, Sun, Poole, Krishnan, Schmid, and
  Isola}]{tian2020makes}
Tian, Y.; Sun, C.; Poole, B.; Krishnan, D.; Schmid, C.; and Isola, P. 2020.
\newblock What makes for good views for contrastive learning?
\newblock \emph{Advances in neural information processing systems}, 33:
  6827--6839.

\bibitem[{Vedantam, Lawrence~Zitnick, and Parikh(2015)}]{vedantam2015cider}
Vedantam, R.; Lawrence~Zitnick, C.; and Parikh, D. 2015.
\newblock Cider: Consensus-based image description evaluation.
\newblock In \emph{Proceedings of the IEEE conference on computer vision and
  pattern recognition}, 4566--4575.

\bibitem[{Velickovic et~al.(2017)Velickovic, Cucurull, Casanova, Romero, Lio,
  Bengio et~al.}]{velickovic2017graph}
Velickovic, P.; Cucurull, G.; Casanova, A.; Romero, A.; Lio, P.; Bengio, Y.;
  et~al. 2017.
\newblock Graph attention networks.
\newblock \emph{stat}, 1050(20): 10--48550.

\bibitem[{Wei et~al.(2022)Wei, Wang, Schuurmans, Bosma, Xia, Chi, Le, Zhou
  et~al.}]{wei2022chain}
Wei, J.; Wang, X.; Schuurmans, D.; Bosma, M.; Xia, F.; Chi, E.; Le, Q.~V.;
  Zhou, D.; et~al. 2022.
\newblock Chain-of-thought prompting elicits reasoning in large language
  models.
\newblock \emph{Advances in Neural Information Processing Systems}, 35:
  24824--24837.

\bibitem[{Wu and Mooney(2019{\natexlab{a}})}]{wu2019faithful}
Wu, J.; and Mooney, R. 2019{\natexlab{a}}.
\newblock Faithful Multimodal Explanation for Visual Question Answering.
\newblock In \emph{Proceedings of the 2019 ACL Workshop BlackboxNLP: Analyzing
  and Interpreting Neural Networks for NLP}, 103--112.

\bibitem[{Wu and Mooney(2019{\natexlab{b}})}]{wu2019self}
Wu, J.; and Mooney, R. 2019{\natexlab{b}}.
\newblock Self-critical reasoning for robust visual question answering.
\newblock \emph{Advances in Neural Information Processing Systems}, 32.

\bibitem[{Xiong, Merity, and Socher(2016)}]{xiong2016dynamic}
Xiong, C.; Merity, S.; and Socher, R. 2016.
\newblock Dynamic memory networks for visual and textual question answering.
\newblock In \emph{International conference on machine learning}, 2397--2406.
  PMLR.

\bibitem[{Yang et~al.(2022)Yang, Li, Hu, Ma, Ding, and Zhang}]{yang2022chunk}
Yang, Q.; Li, Y.; Hu, B.; Ma, L.; Ding, Y.; and Zhang, M. 2022.
\newblock Chunk-aware alignment and lexical constraint for visual entailment
  with natural language explanations.
\newblock In \emph{Proceedings of the 30th ACM International Conference on
  Multimedia}, 3587--3597.

\bibitem[{Yao et~al.(2019)Yao, Pan, Li, and Mei}]{yao2019hierarchy}
Yao, T.; Pan, Y.; Li, Y.; and Mei, T. 2019.
\newblock Hierarchy parsing for image captioning.
\newblock In \emph{Proceedings of the IEEE/CVF international conference on
  computer vision}, 2621--2629.

\bibitem[{Zhang, Zhang, and Xu(2021)}]{zhang2021multi}
Zhang, X.; Zhang, F.; and Xu, C. 2021.
\newblock Multi-level counterfactual contrast for visual commonsense reasoning.
\newblock In \emph{Proceedings of the 29th ACM International Conference on
  Multimedia}, 1793--1802.

\bibitem[{Zhang et~al.(2020)Zhang, Zhao, Lin, He
  et~al.}]{zhang2020counterfactual}
Zhang, Z.; Zhao, Z.; Lin, Z.; He, X.; et~al. 2020.
\newblock Counterfactual contrastive learning for weakly-supervised
  vision-language grounding.
\newblock \emph{Advances in Neural Information Processing Systems}, 33:
  18123--18134.

\end{thebibliography}

\end{document}